\documentclass{article}

\usepackage{microtype}
\usepackage{graphicx}
\usepackage{subfigure}
\usepackage{booktabs} 

\usepackage{hyperref}

\usepackage[accepted]{icml2025}

\usepackage{amsmath}
\usepackage{amssymb}
\usepackage{mathtools}
\usepackage{amsthm}

\usepackage{graphicx}
\usepackage{color}
\usepackage{subcaption}
\usepackage{multirow}
\usepackage{colortbl}
\usepackage{listings}
\usepackage{xcolor}

\usepackage[capitalize,noabbrev]{cleveref}

\theoremstyle{plain}
\newtheorem{theorem}{Theorem}[section]

\theoremstyle{definition}
\newtheorem{definition}[theorem]{Definition}

\theoremstyle{remark}

\usepackage[textsize=tiny]{todonotes}

\icmltitlerunning{Towards the Causal Complete Cause of Multi-Modal Representation Learning}

\begin{document}

\twocolumn[
\icmltitle{Towards the Causal Complete Cause of Multi-Modal Representation Learning}

\begin{icmlauthorlist}
\icmlauthor{Jingyao Wang}{1,2}
\icmlauthor{Siyu Zhao}{2}
\icmlauthor{Wenwen Qiang}{1,2}
\icmlauthor{Jiangmeng Li}{1,2}\\
\icmlauthor{Changwen Zheng}{1,2}
\icmlauthor{Fuchun Sun}{1,3}
\icmlauthor{Hui Xiong}{4}
\end{icmlauthorlist}

\icmlaffiliation{1}{Institute of Software Chinese Academy of Sciences, Beijing, China}
\icmlaffiliation{2}{University of the Chinese Academy of Sciences, Beijing, China}
\icmlaffiliation{3}{Tsinghua University, Beijing, China}
\icmlaffiliation{4}{Thrust of Artificial Intelligence, The Hong Kong University of Science and Technology (Guangzhou), China
Department of Computer Science and Engineering, The Hong Kong University of Science and Technology Hong Kong SAR, China}

\icmlcorrespondingauthor{Wenwen Qiang}{qiangwenwen@iscas.ac.cn}

\icmlkeywords{Machine Learning, ICML}

\vskip 0.3in
]



\printAffiliationsAndNotice{}  

\begin{abstract}
Multi-Modal Learning (MML) aims to learn effective representations across modalities for accurate predictions. Existing methods typically focus on modality consistency and specificity to learn effective representations. However, from a causal perspective, they may lead to representations that contain insufficient and unnecessary information. To address this, we propose that effective MML representations should be causally sufficient and necessary. Considering practical issues like spurious correlations and modality conflicts, we relax the exogeneity and monotonicity assumptions prevalent in prior works and explore the concepts specific to MML, i.e., Causal Complete Cause ($C^3$). We begin by defining $C^3$, which quantifies the probability of representations being causally sufficient and necessary. We then discuss the causal identifiability of $C^3$ and introduce an instrumental variable to support identifying $C^3$ with non-exogeneity and non-monotonicity. Building on this, we conduct the $C^3$ measurement, i.e., $C^3$ risk. We propose a twin network to estimate it through (i) the real-world branch: utilizing the instrumental variable for sufficiency, and (ii) the hypothetical-world branch: applying gradient-based counterfactual modeling for necessity. Theoretical analyses confirm its reliability. Based on these results, we propose $C^3$ Regularization, a plug-and-play method that enforces the causal completeness of the learned representations by minimizing $C^3$ risk. Extensive experiments demonstrate its effectiveness.
\end{abstract}


\section{Introduction}
\label{sec:1}

The initial inspiration for artificial intelligence is to imitate human perceptions which are based on different modalities \cite{mlsurvey1,mlsurvey2}, e.g., sight, sound, and touch. Each modality provides unique information with distinct statistical properties \cite{baltruvsaitis2018multimodal, wang2023amsa}. A fundamental aspect of human perception is the ability to simultaneously integrate data from different modalities to understand the world \cite{mmlsurvey}. Multi-modal learning (MML) \cite{huang2021makes, ge2023improving} has emerged as a promising approach to emulate human sensory perception. It aims to learn good representations from multiple modalities that can achieve accurate decisions.

In this context, a key question arises: what defines a ``good'' MML representation? Existing methods typically define good MML representations from two perspectives: modality consistency and modality specificity \cite{zhang2023generalization,ge2023improving,radford2021learning,fan2024improving,dong2024simmmdg}. The former emphasizes extracting modality-shared semantics related to primary events within the MML task. This type of method \cite{implicit2,xia2024achieving,abdollahzadeh2021revisit} aims to obtain unified representations by mapping features from different modalities into a common embedding space. In contrast, the second perspective believes modality specificity captures the distinct statistical properties of each modality, reflecting different aspects of the primary events. This type of method \cite{dong2024simmmdg,yang2022learning,frost2015domain,zhou2021specificity} decomposes features within each modality into modality-specific and modality-shared components, learning all shared components while applying distance constraints on modality-specific features to enhance diversity.

\begin{figure}
    \centering
    \includegraphics[width=\linewidth]{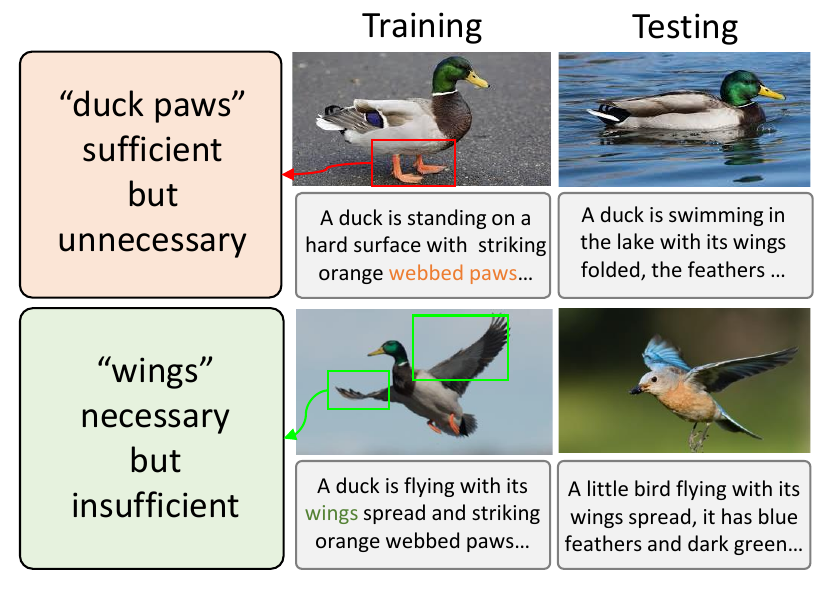}
    \caption{Example of causal sufficiency and necessity in ``duck'' classification task (See \textbf{\Cref{sec:3}} for more analyses).}
    \label{fig:example}
\end{figure}

However, from a causal perspective \cite{ahuja2020invariant,koyama2020invariance}, these methods may result in the learned representations being insufficient or unnecessary. 
Specifically, sufficiency indicates that the use of the representations will establish the label, while necessity indicates that the label becomes incorrect when the representations are absent \cite{pearl2009causality}. \textbf{Figure \ref{fig:example}} provides an example. If the MML model only focuses on causal sufficiency, it will lose important modality-specific semantics, affecting generalization; if the model only focuses on causal necessity, the decisions may be made incorrectly based on the background, affecting discriminability. 
Correspondingly, considering the aforementioned two types of MML methods: (i) those focused on modality consistency may cause the model to overlook important modality-specific semantics; (ii) those focused on modality specificity may incorrectly capture irrelevant information.
Thus, existing MML methods \cite{dong2024simmmdg,explicitly1} without causal constraints may fail to satisfy sufficiency and necessity, affecting model performance. 
The experiments in \textbf{Section \ref{sec:7.2}} further prove this (\textbf{Figure \ref{fig:ex_3}} and \textbf{Table \ref{tab:ex_1}}): (i) the representations learned by existing methods have much lower correlation scores with sufficient and necessary causes than the proposed method with causal constraints; (ii) after constraining the learned representations, the performance of existing methods are significantly improved. The analyses in \textbf{Section \ref{sec:5}} also emphasize the importance of causal sufficiency and necessity.
Thus, inspired by \cite{pearl2009causality,yang2024invariant}, we propose that a good MML representation must be both causally sufficient and necessary, i.e., causally complete. 

Noticeably, in practical MML applications, constraining the causal sufficiency and necessity of learned representations remains challenging.
Specifically, the related discussion in existing works \cite{yang2024invariant,pearl2009causality} is based on the exogeneity and monotonicity assumptions (\textbf{Definition D.1} and \textbf{Definition D.2} in \textbf{Appendix D.3}). Exogeneity refers to the scenario where the influence of the external intervention on the conditional distributions is negligible when the causal representation variable is exogenous relative to the label variable, while monotonicity illustrates the consistent, unidirectional effect on the label of the representation. However, in practice, non-trivial corner cases often arise: the inseparability of MML semantics leads to spurious correlations (\textbf{Figure \ref{fig:scm}}), and cross-modal conflicts combined with high-dimensional nonlinear interactions \cite{huang2021learning,wang2023hacking} undermine monotonicity. These challenges render the conditions of traditional causal sufficiency and necessity discussion inapplicable.

To address this, in this paper, we relax the above assumptions to explore the causal sufficiency and necessity in MML, ensuring the quality of the learned representation. 
Firstly, we propose the definition of causal sufficiency and necessity, i.e., causal complete causes ($C^3$), for MML based on \cite{pearl2009causality}. It reflects the probability that the learned representation is causally complete by estimating the probability of label change after an intervention on the representation, given two conditions of observations. One condition is for sufficiency, and another is for necessity.
Then, we analyze the causal identifiability of $C^3$, which allows us to quantify $C^3$ using the observable data in practice.
Unlike previous studies, we propose an instrumental variable to ensure estimating the segmented effect of \( C^3 \) without spurious correlations, thereby relaxing the assumptions of exogeneity and monotonicity.
Based on this, we propose the measurement of $C^3$ using the twin network, i.e., $C^3$ risk, where a low $C^3$ risk means that the learned representations are causally complete with high confidence. 
The challenge of \( C^3 \) measurement lies in eliminating the spurious correlations in the sufficiency evaluation and modeling the counterfactual data required for the necessity evaluation. The proposed twin network addresses this through (i) the real-world branch: using the proposed instrumental variable to eliminate spurious correlations in the representation, and (ii) the hypothetical-world branch: using provable gradient-based adjustments to model the counterfactuals.
Through theoretical analyses, we prove the reliability of the twin network and provide the performance guarantee of the $C^3$ risk. Based on these theoretical results, we propose Causal Complete Cause Regularization ($C^3$R), a plug-and-play method to learn causal complete representations by constraining their $C^3$ risks.

\textbf{The main contributions are as follows:} (i) We propose the definition, identifiability, and measurement of the causal complete cause ($C^3$) concept for MML without the assumptions of exogeneity and monotonicity. These results provide an effective way, i.e., $C^3$ risk, to estimate the sufficiency and necessity of the learned representation for robust and accurate MML (\textbf{Section \ref{sec:4}}). (ii) We theoretically demonstrate the effectiveness and reliability of the proposed $C^3$ risk. Inspired by this, we propose $C^3$R, which can be applied to any MML model to learn causal complete representations with low $C^3$ risk (\textbf{Sections \ref{sec:5} and \ref{sec:6}}). (iii) Finally, we conduct extensive experiments on various datasets that prove the effectiveness and robustness of $C^3$R (\textbf{Section \ref{sec:7}}).

\section{Problem Analysis}
\label{sec:3}

\paragraph{Problem Settings} 
Let \(\mathcal{X}\), \(\mathcal{Y}\), and \(\mathcal{Z}\) represent the input space, label space, and latent space, respectively. Consider a MML task involving data \((x, y) \in \mathcal{X} \times \mathcal{Y}\), where \(x \in \mathcal{X}\) denotes the sample and \(y \in \mathcal{Y}\) denotes the label. All the data are sampled from the same joint distribution $P_{XY}$ over $\mathcal{X} \times \mathcal{Y}$. The training dataset is defined as $\mathcal{D}_{tr} = \{(x_i, y_i)\}_{i=1}^N$, and the testing dataset is defined as $\mathcal{D}_{te}$. Here, each sample \(x_i = \{x^{(1)}_i, \dots, x^{(K)}_i\}\) contains \(K\) modalities. 
The objective of MML is to obtain a robust model $f_{\theta}=\mathcal{W} \circ g$ that performs well on the unseen test dataset $\mathcal{D}_{te}$. Here, the $g: \mathcal{X} \rightarrow \mathcal{Z}$ is the feature extractor and $\mathcal{W}: \mathcal{Z} \rightarrow \mathcal{Y}$ is the classifier. The specific implementation is illustrated in \textbf{Appendix G}, and the table of notations is shown in \textbf{Table 4}.

From the perspective of data generation \cite{suter2019robustly,hu2022improving}, each sample is considered to be generated using a set of generating factors, e.g., color, shape, etc. In the context of MML, each sample contains generating factors from all modalities. From a causal perspective \cite{pearl2009causality}, these generating factors can be divided into task-related factors, i.e., causal factors \(\text{F}_c\), and task-independent factors, i.e., non-causal factors $\text{F}_{s}$. 
Among them, the causal factors are related to the task label, supporting accurate predictions \cite{deshpande2022deep}.
Thus, the goal of $f_\theta$ can be re-established as learning representations that contain all causal generating factors $\text{F}_c$ from the data without $\text{F}_s$.

\paragraph{Example of Causal Sufficiency and Necessity} 
As shown in \textbf{Figure \ref{fig:example}}, assume the goal of this MML task is to classify the ``duck'', the analysis is conducted under specific task and data conditions. The training data of this task contains three modalities, i.e., image, text, and audio, and all the modalities have the feature ``duck paws''. Then, the models that learn representations based on consistency may capture the features of ``duck paws'', and can establish the label ``duck'' based on ``duck paws''. However, the model may make errors in another ``duck'' scenario where the MML samples do not contain ``duck paws'', e.g., ``duck swims on the lake'' (upper right in \textbf{Figure \ref{fig:example}}). This suggests that the learned representation contains \textit{sufficient but unnecessary} causes: the label ``duck'' can be predicted using the current representation (Line 1 Left), but may not work in another scenario (Line 1 Right).  
Similarly, the models may also learn \textit{necessary but insufficient} representations, e.g., the representations with feature ``wings''. Take the Line 2 of \textbf{Figure \ref{fig:example}} as an example, ``duck'' must have ``wings'', but the samples with ``wings'' may also correspond to another label, e.g., ``bird'' (lower of \textbf{Figure \ref{fig:example}}), leading to incorrect predictions.
Under this context, a sufficient and necessary feature can be ``flat duck bill'' as its presence indicates ``duck'' and every ``duck'' sample includes it. 
Briefly, for \textit{sufficient and necessary} causes, they ensure that the learned representations not only reflect the features targeted the ``duck'' label, but also the features that the ``duck'' label must have. 
Thus, a good representation must have both causal sufficiency and necessity, i.e., causal complete causes, for accurate and robust prediction. This is also the goal we explore in this study. More analyses are provided in \textbf{Appendix D.2}.

\begin{figure}[t]
    \centering
    \includegraphics[width=\linewidth]{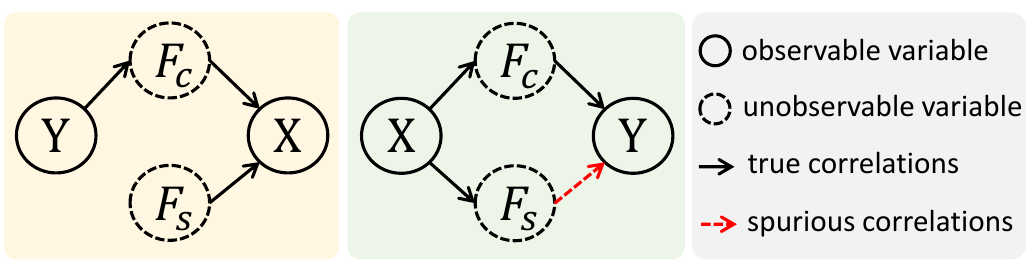}
    \caption{Structural Causal Model (SCM) for MML. \textbf{Left:} causal generating mechanism, \textbf{Right:} the learning process.}
    \label{fig:scm}
\end{figure}

\paragraph{Causal Analysis of MML}
We construct a Structural Causal Model (SCM) for MML based on the causal generating mechanism ~\cite{suter2019robustly,deshpande2022deep} in \textbf{Figure \ref{fig:scm} left}. In this SCM, $\text{Y}$ and $\text{X}$ denote the label variable and corresponding generated data variable in the MML task. $\text{F}_{c}$ and $\text{F}_{s}$ represent the distinct sets of generating factors that are causally and non-causally related to $\text{Y}$. Each generating factor corresponds to a semantic of the MML task, e.g., color, shape, background, modality indicator, etc. Since both $\text{F}_{c}$ and $\text{F}_{s}$ represent high-level knowledge of the data but only $\text{F}_{c}$ is causally related to $\text{Y}$, we could naturally define the MML task label variable $\text{Y}$ as the cause of the $\text{F}_{c}$, i.e., ${\text{Y}} \rightarrow \text{F}_{c}$. There is no connection between $\text{F}_{s}$ and $\text{Y}$.
Following \cite{hu2022causal,deshpande2022deep}, the sample in each modality is generated simultaneously using all the generating factors, including factors caused from the label and unobservable variables, e.g., environmental effects.
The unobservable variable results in the generating factors $\text{F}_s$, which are non-causally related to $\text{Y}$.
Then, we get both $\text{F}_c \rightarrow \text{X}$ and $\text{F}_s \rightarrow \text{X}$. 
Thus, we obtain \textbf{Figure \ref{fig:scm} left}. 
Based on this, we further construct an SCM to discuss the learning process of MML, as shown in \textbf{Figure \ref{fig:scm} right}.
It can be viewed as the inverse process of the causal generating mechanism.
Based on \textbf{Figure \ref{fig:scm} left}, an ideal MML model should only utilize causal generating factors $\text{F}_c$ and be invariant to any intervention on non-causal generating factors $\text{F}_{s}$.
However, in practice, $\text{F}_{c}$ and $\text{F}_{s}$ in the learned representation may be coupled \cite{dong2024simmmdg,li2023decoupled}. This leads to the model potentially learning based on non-causal $\text{F}_s$. There exist spurious correlations between $\text{F}_{s}$ and $\text{Y}$, i.e., we get the additional $\text{F}_{s} \rightarrow \text{Y}$ in \textbf{Figure \ref{fig:scm} right}.

The presence of spurious correlations leads to inaccuracies in MML. Specifically, the learned MML representations may fall into four scenarios: (i) sufficient and necessary: contains all $\text{F}_c$; (ii) necessary but insufficient: contains part of $\text{F}_c$; (iii) sufficient but unnecessary: contains all $\text{F}_c$ with $\text{F}_s$; (iv) insufficient and unnecessary: only contains $\text{F}_s$. 
See \textbf{Appendix D} and \textbf{G} for more analyses. To ensure accurate and robust MML, in this paper, we aim to propose a methodology that constrains the model to learn representations that contain only causally sufficient and necessary information.

\section{Causal Complete Cause}
\label{sec:4}

To access the learning of causal sufficient and necessary representations, in this section, we first provide the definition of causal sufficiency and necessity in MML, i.e., Causal Complete Cause ($C^3$). Next, we discuss the identifiability of $C^3$, which ensures the quantification of $C^3$ through observable data in practice, even in corner cases (non-monotonicity and
non-exogeneity). Finally, we give the measurement of $C^3$, i.e., $C^3$ risk, to measure the probability of whether the learned MML representation is causally complete.

\subsection{Definition of $C^3$}
\label{sec:4.1}
To learn representations of causal variable $\text{F}_{c}$ for MML that is with both causal sufficiency and necessity, based on \cite{pearl2009causality,yang2024invariant}, we introduce the concept of the probability of \textit{Causal Complete Cause} ($C^3$) as follows:

\begin{definition}[Probability of Causal Complete Cause ($C^3$)]\label{def:C3} 
    Denote that the data and corresponding label variables of the given multi-modal data distribution are $\text{X}$ and $\text{Y}$, while the variable of the learned MML representation is $\text{Z}$. Let the specific implementations of representation variable $\text{Z}$ as $c$ and $\bar{c}$, where $c$ denotes the implementation that results in the accurate label prediction $\text{Y}=y$, and $\bar{c}\not=c$ denotes the implementation resulting in $\text{Y}\not=y$. The probability that $\text{Z}$ is the causal complete cause of $\text{Y}$ can be defined as:
\begin{equation}\label{eq:c3}
\begin{split}
    C^3(\text{Z}) :=&\scalebox{0.85}{$\underbrace{P(\text{Y}_{do(\text{Z}=c)}=y\mid \text{Z} = \bar{c}, \text{Y}\not=y)}_{\text{Sufficiency}} P( \text{Z}= \bar{c}, \text{Y}\not=y)$} \\
    &\scalebox{0.85}{$+\underbrace{P(\text{Y}_{do(\text{Z}=\bar{c})}\not=y\mid \text{Z} = c, \text{Y}=y)}_{\text{Necessity}} P( \text{Z} = c, \text{Y}=y)$},
\end{split}
\end{equation}
    where $P(\text{Y}_{do(\text{Z}=c)}=y\mid \text{Z} = \bar{c}, \text{Y}\not=y)$ denotes the probability of $\text{Y}=y$ when force $\text{Z}$ to be another specific implementation $c$ via do-operator $do(\text{Z}=c)$, when giving observations $\text{Z} = \bar{c}$ and $\text{Y}\not=y$ with probability $P( \text{Z}= \bar{c}, \text{Y}\not=y)$. Similarly, the second term corresponds to the case that where the observations are $\text{Z} = c$ and $\text{Y}=y$, the probability of $\text{Y}$ becomes incorrect when force $\text{Z}=\bar{c}$. 
\end{definition}
\textbf{Definition \ref{def:C3}} indicates that when $\text{Z}$ with a high $C^3$ score, it has a high probability of being the causal complete cause of $\text{Y}$. The two terms in Eq.\ref{eq:c3} correspond to sufficiency and necessity, respectively. 
Specifically, the sufficiency term $P(\text{Y}_{do(\text{Z}=c)}=y\mid \text{Z} = \bar{c}, \text{Y}\not=y)$ means that even when \(\text{Z} = \bar{c}\) and \(\text{Y} \neq y\), intervening to set \(\text{Z} = c\) results in a high probability of \(\text{Y} = y\), indicating that \(\text{Z} = c\) has a sufficient causal effect on the occurrence of \(\text{Y} = y\).
Correspondingly, for necessity term $P(\text{Y}_{do(\text{Z}=\bar{c})}\not=y\mid \text{Z} = c, \text{Y}=y)$, even when \(\text{Z} = c\) and \(\text{Y} = y\), intervening to set \(\text{Z} = \bar{c}\) results in \(\text{Y} \neq y\), indicating that \(\text{Z} = c\) is necessary for the occurrence of \(\text{Y} = y\). Under this condition, the learned representation can be divided into sufficient but unnecessary causes, necessary but insufficient causes, sufficient and necessary causes, and insufficient and unnecessary causes (\textbf{Appendix D.1} for detailed analyses).
We aim to constrain the $C^3$ score of the learned representation, i.e., extract sufficient and necessary causes, to achieve robust and accurate MML.

\subsection{Causal Identifiability of $C^3$}
\label{sec:4.2}

Since it is difficult to obtain all samples in the multi-modal data distribution, especially in real systems, e.g., the counterfactual data in the definition of $C^3$ is difficult to obtain \cite{kusner2017counterfactual,morgan2015counterfactuals}, calculating the probability of $C^3$ is still a challenging issue. To access the calculation of $C^3$ based on observable MML data, we discuss the causal identifiability of $C^3$ in this section.

Identifiability refers to the ability to uniquely infer causal effects from observable data under given assumptions \cite{pearl2009causality}. For MML settings, it means the causal factors within the representation $Z$ can uniquely determine the label \( Y \) from the sample \( X \). To ensure reliable estimation, we constrain the \( C^3 \) score of the learned representation \( Z \) extracted from \( X \), ensuring its identifiability. 
In previous studies \citep{tian2002general,yang2024invariant}, identifying causal probabilities typically assumes that statistical data are derived under exogeneity and monotonicity conditions (\textbf{Definition D.1} and \textbf{Definition D.2} in \textbf{Appendix D.3}). Exogeneity refers to where the external intervention's impact on the conditional distribution $P(\text{Y}|\text{Z})$ is negligible when the variable \(\text{Z}\) is exogenous to \(\text{Y}\). 
This ensures the learned \text{Z} to recognize $\text{F}_c$ without being affected by $\text{F}_s$.
Monotonicity, on the other hand, describes the consistent, unidirectional effect of the representation \(\text{Z}\) on \(\text{Y}\).
This ensures that the impact of the learned $\text{Z}$ on the labeling decision will not be reversed as other factors change.
However, in practice, non-trivial cases often arise: the inseparability of generating factors in the latent space leads to spurious correlations, i.e., the learned representations $\text{Z}$ contain non-causal $\text{F}_s$ (\textbf{Figure \ref{fig:scm}}); Meanwhile, modality conflicts and the high-dimensional nonlinear interactions \citep{wang2023hacking,huang2021learning} undermine monotonicity. These challenges render traditional identifiability conditions inapplicable.
To address this, we relax the above assumptions and explore the identifiability of \( C^3 \), assessing the calculation of the $C^3$ score.

Consider the effects of spurious correlations, i.e., \( P(\text{Y}_{do(\text{Z}=c)}) \neq P(\text{Y} \mid \text{Z}=c) \), we first relax the exogeneity assumption (with definition and analysis in \textbf{Appendix D.3}). 
Based on \textbf{Definition \ref{def:C3}}, we extend Theorem 9.2.15 in \cite{pearl2009causality} to MML for $C^3$, obtaining:
\begin{theorem}[Causal Identifiability under Non-Exogeneity]
\label{theorem:identifiability_1}
Given the MML model \( f_\theta \), where the label variable \(\text{Y}\) is influenced by the causal factors \(\text{F}_{c}\) and non-causal factors $\text{F}_{s}$. Assume $f_\theta$ satisfy the Positive Markovian assumption \cite{tian2002general}, the probabilities of $C^3$ ($P(\text{Y}_{do(\text{Z} = c)} = y, \text{Y}_{do(\text{Z} = \bar{c})} \neq y)$) is identifiable and can be estimated via:
\begin{equation}
    C^3(\text{Z}) = P(\text{Y}_{do(\text{Z} = c)}) - P(\text{Y}_{do(\text{Z} = \bar{c})})
\end{equation}
where $Y$ is monotonic relative to $\text{Z}$ with $\text{Y}_{do(\text{Z}=c)}=\bar{y} \wedge \text{Y}_{do(\text{Z}=c)}=y$ is false or $\text{Y}_{do(\text{Z}=\bar{c})}=y \wedge \text{Y}_{do(\text{Z}=c)}=\bar{y}$ is false, and the model $f_\theta$ satisfies local invertibility.
\end{theorem}
\textbf{Theorem \ref{theorem:identifiability_1}} establishes that, under the monotonicity condition, \( C^3 \) can be estimated using observable multi-modal data, thereby quantifying \( C^3 \). 
The local invertibility (Proposition D.3) states that the model can uniquely recover the distribution of $s$ from the conditional distribution of $Y$ given its parents $\mathrm{Pa}(Y)$.
Specifically, by modeling the effects of spurious correlations (\( \text{F}_s \) on $\text{Y}$) as non-causal pathways in the SCM, interventions with the \( do \)-operator, such as \( P(\text{Y}_{do(\text{Z} = c)}) \) and \( P(\text{Y}_{do(\text{Z} = \bar{c})}) \), facilitates reliable causal effect estimation. For counterfactual terms ($\text{Z}=\bar{c}$) involved in the necessity probability, the model \( f_\theta \) imposes locally reversible constraints to ensure identifiability, as discussed in \cite{galles1998axiomatic}.
Under these conditions, the probability of causal sufficiency \( C^3_{su} \) and the probability of causal necessity \( C^3_{ne} \) can be estimated as follows: 
\begin{equation}\label{eq:c3_su}
    \scalebox{0.97}{$C^3_{su}(\text{Z}) = \frac{P(\text{Y}_{do(\text{Z}= c)}) - P(\text{Y}_{do(\text{Z} = \bar{c})})}{1 - P(\text{Z} = c)}$}
\end{equation}
\begin{equation}\label{eq:c3_ne}
    \scalebox{0.97}{$C^3_{ne}(\text{Z}) = \frac{P(\text{Y}_{do(\text{Z} = c)}) - P(\text{Y}_{do(\text{Z} = \bar{c})})}{P(\text{Z} = c)}$}
\end{equation}

Next, according to the second paragraph of \textbf{Subsection \ref{sec:4.2}}, considering the cross-modal conflicts with high-dimensional nonlinear interactions in practice, we relax the assumptions of monotonicity. The key idea lies in introducing an instrumental variable \( V \) \cite{caner2004instrumental} to estimate the piecewise effects of conditional distributions within the $C^3$ definition. Meanwhile, $V$ satisfies \( V \perp\!\!\!\perp \text{F}_s \) with a controllable impact on \(\text{F}_c\), i.e.,
$V \to \text{F}_c \to \text{Y}$. Then, we have:
\begin{theorem}[Causal Identifiability under Non-Monotonicity and Non-Exogeneity]
\label{theorem:identifiability_2}
    Given \( f_\theta \) that learns the causal effect $\text{Z}\to \text{Y}$. The representation variable $\text{Z} = \Xi_c\text{F}_c+\Xi_s \text{F}_s$ where $\Xi_{(\cdot)}$ is the weight matrix. 
    Introducing an instrumental variable \( V \) satisfying \( P(\text{Y} \mid \text{Z}, V) \approx P(\text{Y} \mid \text{Z}) \), we have:  
\begin{equation}\label{eq:c3_non_assumptions}
\begin{split}
    C^3(\text{Z}) & = \int_v \big[P(\text{Y}=y \mid \text{Z} = c, V = v) \\
    & - P(\text{Y}=y \mid \text{Z} = \bar{c}, V = v)\big] P(V = v) \, \text{d}v.
\end{split}
\end{equation}
\end{theorem}
\textbf{Theorem \ref{theorem:identifiability_2}} posits that the estimation of $C^3$ is achievable without the dependence on exogeneity and monotonicity, enabling the quantification of $C^3$ in the absence of counterfactual data. The detailed discussion is provided in \textbf{Appendix D.3}, and the proofs of the above theorems are provided in \textbf{Appendix A.1 and A.2}, respectively. 
According to this theorem, the practical estimation of \( C^3 \) hinges on how the instrumental variable \( V \) is modeled and how the sufficiency and necessity terms of \( C^3 \) are constrained under \( V \), also the main focus in the next subsection, i.e., measurement of $C^3$.

\subsection{Measurement of $C^3$}
\label{sec:4.3}

Based on \textbf{Definition \ref{def:C3}} and \textbf{Theorem \ref{theorem:identifiability_2}}, we provide the measurement of $C^3$ in this section, i.e., $C^3$ risk, to estimate the $C^3$ score of the representation distribution $P(\text{Z}|\text{X}=x)$ inferred from $\text{X}$ on the multi-modal data distribution $P_{XY}$. When the learned representation obtains less necessary and sufficient information, the $C^3$ risk will be higher. 
Specifically, we first model the instrumental variable \(V\) based on \textbf{Theorem \ref{theorem:identifiability_2}}. It aims to make the MML model learn representations $\text{Z}$ that only contain the causal factors $\text{F}_c$ for decision-making while ensuring the identifiability of the \(C^3\) score assessment (\textbf{Theorem \ref{theorem:variable_Z}}). Next, based on \textbf{Definition \ref{def:C3}}, we provide the $C^3$ measurement of the learned $\text{Z}$, i.e., $ C^3$ risk (\textbf{Theorem \ref{theorem:c3r}}). We employ a twin network (\textbf{Figure 5}) with a real-world branch (extract causal representation using the proposed instrumental variable) and a hypothetical-world branch (modeling counterfactual data with gradient-based adjustments), thereby modeling the sufficiency and necessity terms. \textbf{Appendix D.4} establishes the reliability of the proposed twin network.

Firstly, as shown in \textbf{Section \ref{sec:3}}, we establish the model $f_\theta$ using the feature extractor $g$ and the linear classifier $\mathcal{W}: \mathbb{R}^d \rightarrow \mathcal{Y}$ on the overall representation $\text{Z}=g(\text{X})$ ($\text{Z} = \Xi_c\text{F}_c+\Xi_s \text{F}_s$) to obtain the label $\text{Y} = {\rm sign} (\mathcal{W}^{\top}g(\text{X}))$. To infer $\text{F}_{c}$ from observable data $x\sim \mathcal{X}$, we use the instrumental variable $V$. Here, we focus on how \( V \) is modeled to constrain \( \text{Z} \) only contains $\text{F}_c$ for decision-making while satisfying the conditions in \textbf{Theorem \ref{theorem:identifiability_2}}. We propose:
\begin{theorem}[Instrumental Variable \(V\) in MML]
\label{theorem:variable_Z}
For multimodal input \( X \in \mathbb{R}^{K \times d} \), representing \( K \) modalities each with a feature dimension of \( d \), instrumental variable \( V \) is introduced to extract causal generating factors \( \text{F}_c \) while reducing the influence of non-causal factors \( \text{F}_s \). \( V \) is generated from \( X \) using self-attention mechanism with:
\begin{equation}\label{eq:z}
\begin{split}
    V = \sum_{j=1}^K \frac{\exp(s_{ij})}{\sum_{k=1}^K \exp(s_{ik})} X_j W_V,\\
    \text{where}\quad s_{ij} = \frac{q_i^\top k_j}{\sqrt{d}} - \alpha \|q_i - k_j\|^2.
\end{split}
\end{equation}
where \( q_i = X_i W_Q \) and \( k_j = X_j W_K \) are the query and key of \( i \)-th and \( j \)-th modalities' features, \( W_Q, W_K, W_V \in \mathbb{R}^{d \times d} \) are linear projection matrices, mapping query, key and value respectively. Then, $V$ satisfies (i) $P(\text{Z} \mid V) \neq P(\text{Z})$; (ii) $P(\text{F}_s \mid V) = P(\text{F}_s)$; and (iii) $P(\text{Y} \mid \text{Z}, V) = P(\text{Y} \mid \text{F}_c)$.
\end{theorem}
In \textbf{Theorem \ref{theorem:variable_Z}}, we propose using the self-attention mechanism to model the instrumental variable $V$ for $C^3$. Specifically, 
the alignment score $s_{ij}$ is to capture inter-modal interactions for labels. Then, by strengthening the correlation between \( V \) and \( \text{Z} \), the learned representation \( \text{Z} \) is guided to approach \( \text{F}_c \) while weakening the correlation with \( \text{F}_s \). Furthermore, consider inter-modal misalignment, \( \|q_i - k_j\|^2 \) with weight $\alpha$ is introduced as a distance-based adjustment to diminish the impact of modalities with large feature distances on attention weights. 
It satisfies three conditions: (i) Relevance: \( V \) manipulate \( \text{F}_c \); (ii)  Independence: \( V \) is conditionally independent from \( \text{F}_s \); and (iii) Exclusion: the influence of \( V \) on \( \text{Y} \) is completely indirectly realized through \( \text{F}_c \). The detailed proofs are provided in \textbf{Appendix A.3}.

Next, based on these results, we provide the measurement of $C^3$, i.e., $C^3$ risk. 
Specifically, we first utilize \(V\) to constrain \(f_\theta\), effectively distinguishing \(\text{F}_c\) from \(\text{F}_s\) and ensuring the identifiability of \(C^3\) based on \textbf{Theorem \ref{theorem:variable_Z}}. Obtaining $\text{Z}_c$ that only contains $\text{F}_c$, we then define sufficiency risk (when \(\text{Z} = c\), the label is not \(y\)) and necessity risk (when \(\text{Z} = \bar{c}\), the label is \(y\)) to construct the \(C^3\) risk based on \textbf{Definition \ref{def:C3}} and Eq.\ref{eq:c3_non_assumptions}.
The key challenge lies in the counterfactual data, i.e., \(\overline{\text{Z}}_c: \text{Z} = \bar{c}\), is difficult to obtain \cite{kusner2017counterfactual,morgan2015counterfactuals}. 
To address this, we propose a twin network comprising: (i) the real-world branch to obtain \(\widehat{\text{Z}}_c\) with $V$, and (ii) the hypothetical-world branch to obtain \(\overline{\text{Z}}_c\) under a provable gradient-based adjustment, ensuring \(\overline{\text{Z}}_c\) stay close to the original observable distribution (\textbf{Appendix D.4}). These branches share network structures and parameters, maintaining a mirrored correspondence that enforces causal consistency \cite{pearl2009causality}. Thus, we get:
\begin{theorem}[$C^3$ Risk]\label{theorem:c3r}
    Given \(N\) observable samples \(\{(x_i,\,y_i)\}_{i=1}^N\) drawn from the multi-modal distribution \(P_{XY}\), and the model $f_\theta$ with feature extractor \(g(\cdot)\) and linear classifier $\mathcal{W}$, we define the following twin-network:
\begin{equation}\label{eq:twin}
\begin{split}
&\widehat{\text{Z}}_{c,i} = g^*(x_i,v_i),
\quad
\overline{\text{Z}}_{c,i} = \widehat{\text{Z}}_{c,i} + \Delta_i, \\
&\text{where}\quad \Delta_i = -\nabla_{\widehat{\text{Z}}_{c,i}}\ell\Bigl(\sigma(\mathcal{W}^\top \widehat{\text{Z}}_{c,i}),\,y_i\Bigr),\\
&\text{s.t.}\quad \mu_{\text{KL}}(\overline{\text{Z}}_{c,i}, \widehat{\text{Z}}_{c,i}) \leq \varepsilon, \quad \forall i,
\end{split}
\end{equation}
where $g^*(x_i,v_i)$ is the calibrated $g$ with $v_i$ using Kullback–Leibler divergence $\mu_{\text{KL}}(\cdot)$, i.e., with loss $\mathcal{L}_{v}=\mathbb{E}_{(x,y)\sim P_{XY}} \mathbb{E}_{v_i\in V}\mu_{\text{KL}}(g(x),v_i)$, to estimate the causal representation \(\widehat{\text{Z}}_{c,i}\) in the real-world branch, $\Delta_i$ arises from the gradient-based intervention with $ v_i \sim V $ for the hypothetical world. If $f_\theta$ converges to an optimal solution under a Monte Carlo approximation of \(V\), then by using:
\begin{equation}\label{eq:c3r}
\begin{split}
\scalebox{0.85}{$\widehat{R}^{C^3}
=
\frac{1}{N} \sum_{i=1}^N\Big[ 
  \rho[\sigma(\mathcal{W}^\top \widehat{\text{Z}}_{c,i})\neq y_i]+
  \rho[\sigma(\mathcal{W}^\top \overline{\text{Z}}_{c,i})= y_i]
\Big],$}
\end{split}
\end{equation}
the discrepancy between \(\widehat{R}_{N}^{C^3}\) and the ideal causal complete quantity approaches zero with high probability.
Here, $\sigma $ denotes the signum function, $\rho (\cdot)$ denotes an indicator function which is equal to 1 if the condition in $\rho (\cdot)$ is true.
\end{theorem}
\textbf{Theorem \ref{theorem:c3r}} presents a method to measure the causal completeness of learned representations based on observable data. It leverages instrumental variables \(V\) and a twin network to address the issues of spurious correlations in MML and the unavailability of counterfactual data in practice. The detailed proofs and analysis are provided in \textbf{Appendix A.4}.

\section{Performance Guarantee with $C^3$ Risk}
\label{sec:5}

In this section, we conduct theoretical analyses to establish the connection between the proposed $C^3$ risk and MML generalization performance, proving its effectiveness. 

Before discussing generalization, we first illustrate the empirical and expectation risks of the MML model $f_\theta$ on training and test data, i.e., $\widehat{R}^{C^3}(f_\theta)$ (Eq.\ref{eq:c3r}) and ${R}^{C^3}(f_\theta)=\mathbb{E}_{(x,y)\sim P_{XY}}\rho[\sigma(\mathcal{W}^\top \widehat{\text{F}}_{c,i})\neq y_i]+\rho[\sigma(\mathcal{W}^\top \overline{\text{F}}_{c,i})= y_i]$. Then, we provide the following performance guarantee:

\begin{theorem}[Performance Guarantee via $C^3$]\label{theorem:theo_1}
    Denote the hypothesis set $\mathcal{H}$ of linear classifier $\mathcal{W} \in \mathcal{H} \subseteq \{h: \mathbb{R}^d \rightarrow \mathcal{Y}\}$, let $H = \text{\emph{Pdim}}(\{\ell_h : h\in \mathcal{H}\})$. Then with probability at least $1-\delta$, and training samples that sampling from empirical distribution $\mathcal{D}_{tr}(x, y)$, we have
    \begin{equation}
        {R}^{C^3}(f_\theta) \leq \widehat{R}^{C^3}(f_\theta) +M \mathfrak{R}(\mathcal{H})+\sqrt{\frac{ln(1/\delta)}{2H}}.
    \end{equation}
    Here, $\mathfrak{R}(H)$ is the Rademacher complexity of the hypothesis class $H$, and $M$ is a finite constant.
\end{theorem}

\textbf{Theorem \ref{theorem:theo_1}} derives the upper bound on the generalization error of $C^3$ risk, relating its empirical error on the training data to its performance on unseen samples. Specifically, the generalization error is bounded by the sum of the empirical error, the Rademacher complexity of the hypothesis class of the linear classifier, and a confidence term that controls the probability of exceeding the bound. This result provides a theoretical performance guarantee for optimizing MML models with $C^3$ risk. See \textbf{Appendix A.5} for detailed proofs.

\begin{table*}[t]
\caption{Performance comparison when 50\% samples are corrupted with Gaussian noise, i.e., zero mean with the variance of $N$. ``(N, Avg.)'' and ``(N, Worst.)'' denotes the average and worst-case accuracy. The best results are highlighted in \textbf{bold}.}
\vspace{-0.13in}
\setlength\tabcolsep{1pt}
\begin{center}
\resizebox{1.0\linewidth}{!}{
\begin{tabular}{l|cccc|cccc|cccc|cccc}
\toprule
    \multirow{2}{*}{Method} & \multicolumn{4}{c|}{NYU Depth V2}  & \multicolumn{4}{c|}{SUN RGB-D}                     & \multicolumn{4}{c|}{FOOD 101}  & \multicolumn{4}{c}{MVSA}     \\    & (0,Avg.) & (0,Worst.) & (10,Avg.) & (10,Worst.) 
    & (0,Avg.) & (0,Worst.) & (10,Avg.) & (10,Worst.) 
    & (0,Avg.) & (0,Worst.) & (10,Avg.) & (10,Worst.) 
    & (0,Avg.) & (0,Worst.) & (10,Avg.) & (10,Worst.)  \\ 
\midrule
    CLIP \cite{sun2023eva} & 69.32 & 68.29 & 51.67 & 48.54 & 56.24 & 54.73 & 35.65 & 32.76 & 85.24 & 84.20 & 52.12 & 49.31 & 62.48 & 61.22 & 31.64 & 28.27 \\
    ALIGN \cite{jia2021scaling} & 66.43 & 64.33 & 45.24 & 42.42 & 57.32 & 56.26 & 38.43 & 35.13 & 86.14 & 85.00 & 53.21 & 50.85 & 63.25 & 62.69 & 30.55 & 26.44 \\
    MaPLe \cite{khattak2023maple} & 71.26 & 69.27 & 52.98 & 48.73 & 62.44 & 61.76 & 34.51 & 30.29 & 90.40 & 86.28 & 53.16 & 40.21 & 77.43 & 75.36 & 43.72 & 38.82 \\
    CoOp \cite{jia2022visual} & 67.48 & 66.94 & 49.43 & 45.62 & 58.36 & 56.31 & 39.67 & 35.43 & 88.33 & 85.10 & 55.24 & 51.01 & 74.26 & 73.61 & 42.58 & 37.29 \\
    VPT \cite{jia2022visual} & 62.16 & 61.21 & 41.05 & 37.81 & 54.72 & 53.92 & 33.48 & 29.81 & 83.89 & 82.00 & 51.44 & 49.01 & 65.87 & 64.98 & 32.79 & 29.21 \\
    Late fusion \cite{wang2016learning} & 69.14 & 68.35 & 51.99 & 44.95 & 62.09 & 60.55 & 47.33 & 44.60 & 90.69 & 90.58 & 58.00 & 55.77 & 76.88 & 74.76 & 55.16 & 47.78 \\
    ConcatMML \cite{zhang2021modality} & 70.30 & 69.42 & 53.20 & 47.71 & 61.90 & 61.19 & 45.64 & 42.95 & 89.43 & 88.79 & 56.02 & 54.33 & 75.42 & 75.33 & 53.42 & 50.47 \\
    AlignMML \cite{wang2016learning} & 70.31 & 68.50 & 51.74 & 44.19 & 61.12 & 60.12 & 44.19 & 38.12 & 88.26 & 88.11 & 55.47 & 52.76 & 74.91 & 72.97 & 52.71 & 47.03 \\
    ConcatBow \cite{explicitly3} & 49.64 & 48.66 & 31.43 & 29.87 & 41.25 & 40.54 & 26.76 & 24.27 & 70.77 & 70.68 & 35.68 & 34.92 & 64.09 & 62.04 & 45.40 & 40.95  \\
    ConcatBERT \cite{explicitly3} & 70.56 & 69.83 & 44.52 & 43.29 & 59.76 & 58.92 & 45.85 & 41.76 & 88.20 & 87.81 & 49.86 & 47.79 & 65.59 & 64.74 & 46.12 & 41.81  \\
    MMTM \cite{joze2020mmtm} & 71.04 & 70.18 & 52.28 & 46.18 & 61.72 & 60.94 & 46.03 & 44.28 & 89.75 & 89.43 & 57.91 & 54.98 & 74.24 & 73.55 & 54.63 & 49.72 \\
    TMC \cite{han2020trusted} & 71.06 & 69.57 & 53.36 & 49.23 & 60.68 & 60.31 & 45.66 & 41.60 & 89.86 & 89.80 & 61.37 & 61.10 & 74.88 & 71.10 & 60.36 & 53.37 \\
    LCKD \cite{wang2023learnable} & 68.01 & 66.15 & 42.31 & 40.56 & 56.43 & 56.32 & 43.21 & 42.43 & 85.32 & 84.26 & 47.43 & 44.22 & 62.44 & 62.27 & 43.52 & 38.63 \\
    UniCODE \cite{xia2024achieving} & 70.12 & 68.74 & 44.78 & 42.79 & 59.21 & 58.55 & 46.32 & 42.21 & 88.39 & 87.21 & 51.28 & 47.95 & 66.97 & 65.94 & 48.34 & 42.95 \\
    SimMMDG \cite{dong2024simmmdg} & 71.34 & 70.29 & 45.67 & 44.83 & 60.54 & 60.31 & 47.86 & 45.79 & 89.57 & 88.43 & 52.55 & 50.31 & 67.08 & 66.35 & 49.52 & 44.01 \\
    MMBT \cite{kiela2019supervised} & 67.00 & 65.84 & 49.59 & 47.24 & 56.91 & 56.18 & 43.28 & 39.46 & 91.52 & 91.38 & 56.75 & 56.21 & 78.50 & 78.04 & 55.35 & 52.22  \\
    QMF \cite{explicitly3} & 70.09 & 68.81 & 55.60 & 51.07 & 62.09 & 61.30 & 48.58 & 47.50 & 92.92 & 92.72 & 62.21 & 61.76 & 78.07 & 76.30 & 61.28 & 57.61  \\
\midrule
    \rowcolor{orange!10} CLIP+$C^3$R & 76.54 & \textbf{75.12} & 56.73 & 52.90 & 62.31 & 58.71 & 41.59 & 37.52 & 92.93 & 91.80 & 59.77 & 57.54 & 69.61 & 68.64 & 39.58 & 35.89 \\
    \rowcolor{orange!10} MaPLe+$C^3$R & 77.07 & 74.45 & 58.94 & 55.95 & 66.21 & 65.51 & 40.12 & 37.34 & 94.38 & 93.51 & 60.63 & 46.07 & 81.19 & 81.51 & 49.32 & 45.98 \\
    \rowcolor{orange!10}Late fusion+$C^3$R & 73.26 & 71.62 & 57.21 & 50.98 & 64.84 & 63.25 & \textbf{53.35} & 50.43 & 94.09 & 92.24 & 65.27 & 59.02 & \textbf{83.77} & 79.79 & 62.14 & 52.50 \\
    \rowcolor{orange!10}LCKD+$C^3$R & 77.14 & 75.12 & 50.11 & 47.98 & 60.97 & 60.14 & 47.23 & 46.21 & 90.89 & 90.14 & 54.48 & 51.16 & 66.78 & 65.67 & 49.28 & 42.84 \\
    \rowcolor{orange!10}SimMMDG+$C^3$R & 75.32 & 74.61 & 49.99 & 47.22 & 65.50 & 64.58 & 52.69 & \textbf{51.70} & 92.24 & 91.14 & 57.32 & 53.56 & 73.62 & 71.01 & 51.65 & 51.07 \\
    \rowcolor{orange!10}MMBT+$C^3$R & 73.74 & 71.82 & 54.35 & 52.57 & 61.47 & 59.99 & 48.42 & 46.07 & 94.25 & 93.90 & 60.41 & 60.11 & 82.76 & \textbf{81.64} & 62.12 & 58.93 \\
    \rowcolor{orange!10}QMF+$C^3$R & \textbf{77.58} & 74.95 & \textbf{59.72} & \textbf{59.18} & \textbf{67.35} & \textbf{65.84} & 52.26 & 51.28 & \textbf{94.87} & \textbf{93.79} & \textbf{66.45} & \textbf{63.69} & 83.13 & 81.98 & \textbf{66.66} & \textbf{64.51} \\
\bottomrule
\end{tabular}
}\end{center}
\label{tab:ex_1}
\end{table*}

\begin{table*}[t]
\caption{Performance with missing modalities on BraTS. The brackets ``()'' indicate the effect changes after introducing $C^3$R. ``$\bullet$'' and ``$\circ$'' indicate the availability and absence of the modality for testing. The best results are highlighted in \textbf{bold}.}
\vspace{-0.13in}
\setlength\tabcolsep{1pt}
\begin{center}
\resizebox{1.0\linewidth}{!}{
\begin{tabular}{cccc|cccccc|cccccc|cccccc}
\toprule
    \multicolumn{4}{c|}{Modalities}               & \multicolumn{6}{c|}{Enhancing Tumour}                     & \multicolumn{6}{c|}{Tumour Core}                          & \multicolumn{6}{c}{Whole Tumour}                         \\ 
\midrule
    Fl        & T1        & T1c       & T2  
    & HMIS  & HVED  & RSeg  & mmFm  & LCKD & \cellcolor{orange!10}LCKD+$C^3$R          
    & HMIS  & HVED  & RSeg  & mmFm  & LCKD & \cellcolor{orange!10}LCKD+$C^3$R          
    & HMIS  & HVED  & RSeg  & mmFm  & LCKD & \cellcolor{orange!10}LCKD+$C^3$R \\ 
\midrule
$\bullet$ & $\circ$  & $\circ$   & $\circ$   
    & 11.78 & 23.80 & 25.69 & 39.33 & 45.48 & \cellcolor{orange!10}49.81 (\textcolor{red}{+4.33}) 
    & 26.06 & 57.90 & 53.57 & 61.21 & 72.01 & \cellcolor{orange!10}76.65 (\textcolor{red}{+4.64})        
    & 52.48 & 84.39 & 85.69 & 86.10 & 89.45 & \cellcolor{orange!10}91.62 (\textcolor{red}{+2.17}) \\
$\circ$   & $\bullet$ & $\circ$   & $\circ$   
    & 10.16 & 8.60  & 17.29 & 32.53 & 43.22 & \cellcolor{orange!10}49.13 (\textcolor{red}{+6.01})
    & 37.39 & 33.90 & 47.90 & 56.55 & 66.58 & \cellcolor{orange!10}72.18 (\textcolor{red}{+5.60})
    & 57.62 & 49.51 & 70.11 & 67.52 & 76.48 & \cellcolor{orange!10}82.39 (\textcolor{red}{+5.91}) \\
$\circ$   & $\circ$   & $\bullet$ & $\circ$   
    & 62.02 & 57.64 & 67.07 & 72.60 & 75.65 & \cellcolor{orange!10}80.50 (\textcolor{red}{+4.85})
    & 65.29 & 59.59 & 76.83 & 75.41 & 83.02 & \cellcolor{orange!10}88.06 (\textcolor{red}{+5.04})
    & 61.53 & 53.62 & 73.31 & 72.22 & 77.23 & \cellcolor{orange!10}81.93 (\textcolor{red}{+4.70})\\
$\circ$   & $\circ$   & $\circ$   & $\bullet$ 
    & 25.63 & 22.82 & 28.97 & 43.05 & 47.19 & \cellcolor{orange!10}54.13 (\textcolor{red}{+6.94})
    & 57.20 & 54.67 & 57.49 & 64.20 & 70.17 & \cellcolor{orange!10}77.32 (\textcolor{red}{+7.15})         
    & 80.96 & 79.83 & 82.24 & 81.15 & 84.37 & \cellcolor{orange!10}90.78 (\textcolor{red}{+6.41})\\
$\bullet$ & $\bullet$ & $\circ$   & $\circ$   
    & 10.71 & 27.96 & 32.13 & 42.96 & 48.30 & \cellcolor{orange!10}54.16 (\textcolor{red}{+4.86})
    & 41.12 & 61.14 & 60.68 & 65.91 & 74.58 & \cellcolor{orange!10}79.83 (\textcolor{red}{+5.25})
    & 64.62 & 85.71 & 88.24 & 87.06 & 89.97 & \cellcolor{orange!10}93.63 (\textcolor{red}{+3.66})\\
$\bullet$ & $\circ$   & $\bullet$ & $\circ$   
    & 66.10 & 68.36 & 70.30 & 75.07 & 78.75 & \cellcolor{orange!10}82.98 (\textcolor{red}{+4.23})
    & 71.49 & 75.07 & 80.62 & 77.88 & 85.67 & \cellcolor{orange!10}89.74 (\textcolor{red}{+4.07})
    & 68.99 & 85.93 & 88.51 & 87.30 & 90.47 & \cellcolor{orange!10}93.91 (\textcolor{red}{+3.44})\\
$\bullet$ & $\circ$   & $\circ$   & $\bullet$ 
    & 30.22 & 32.31 & 33.84 & 47.52 & 49.01 & \cellcolor{orange!10}56.12 (\textcolor{red}{+7.11})
    & 57.68 & 62.70 & 61.16 & 69.75 & 75.41 & \cellcolor{orange!10}82.57 (\textcolor{red}{+7.16})
    & 82.95 & 87.58 & 88.28 & 87.59 & 90.39 & \cellcolor{orange!10}95.48 (\textcolor{red}{+5.09})\\
$\circ$   & $\bullet$ & $\bullet$ & $\circ$   
    & 66.22 & 61.11 & 69.06 & 74.04 & 76.09 & \cellcolor{orange!10}81.76 (\textcolor{red}{+5.67})
    & 72.46 & 67.55 & 78.72 & 78.59 & 82.49 & \cellcolor{orange!10}88.32 (\textcolor{red}{+5.83})
    & 68.47 & 64.22 & 77.18 & 74.42 & 80.10 & \cellcolor{orange!10}87.03 (\textcolor{red}{+6.96})\\
$\circ$   & $\bullet$ & $\circ$   & $\bullet$ 
    & 32.39 & 24.29 & 32.01 & 44.99 & 50.09 & \cellcolor{orange!10}56.03 (\textcolor{red}{+5.94})
    & 60.92 & 56.26 & 62.19 & 69.42 & 72.75 & \cellcolor{orange!10}78.78 (\textcolor{red}{+6.03})
    & 82.41 & 81.56 & 84.78 & 82.20 & 86.05 & \cellcolor{orange!10}92.33 (\textcolor{red}{+6.28})\\
$\circ$   & $\circ$   & $\bullet$ & $\bullet$ 
    & 67.83 & 67.83 & 69.71 & 74.51 & 76.01 & \cellcolor{orange!10}83.97 (\textcolor{red}{+7.96})
    & 76.64 & 73.92 & 80.20 & 78.61 & 84.85 & \cellcolor{orange!10}93.57 (\textcolor{red}{+8.72})
    & 82.48 & 81.32 & 85.19 & 82.99 & 86.49 & \cellcolor{orange!10}94.40 (\textcolor{red}{+7.91})\\
$\bullet$ & $\bullet$ & $\bullet$ & $\circ$   
    & 68.54 & 68.60 & 70.78 & 75.47 & 77.78 & \cellcolor{orange!10}82.94 (\textcolor{red}{+5.06})
    & 76.01 & 77.05 & 81.06 & 79.80 & 85.24 & \cellcolor{orange!10}90.46 (\textcolor{red}{+5.22})
    & 72.31 & 86.72 & 88.73 & 87.33 & 90.50 & \cellcolor{orange!10}95.51 (\textcolor{red}{+5.01})\\
$\bullet$ & $\bullet$ & $\circ$   & $\bullet$ 
    & 31.07 & 32.34 & 36.41 & 47.70 & 49.96 & \cellcolor{orange!10}56.25 (\textcolor{red}{+6.29})
    & 60.32 & 63.14 & 64.38 & 71.52 & 76.68 & \cellcolor{orange!10}82.69 (\textcolor{red}{+6.01})
    & 83.43 & 88.07 & 88.81 & 87.75 & 90.46 & \cellcolor{orange!10}96.23 (\textcolor{red}{+5.77})\\
$\bullet$ & $\circ$   & $\bullet$ & $\bullet$ 
    & 68.72 & 68.93 & 70.88 & 75.67 & 77.48 & \cellcolor{orange!10}83.90 (\textcolor{red}{+6.42})
    & 77.53 & 76.75 & 80.72 & 79.55 & 85.56 & \cellcolor{orange!10}92.39 (\textcolor{red}{+6.83})
    & 83.85 & 88.09 & 89.27 & 88.14 & 90.90 & \cellcolor{orange!10}96.78 (\textcolor{red}{+5.88})\\
$\circ$   & $\bullet$ & $\bullet$ & $\bullet$ 
    & 69.92 & 67.75 & 70.10 & 74.75 & 77.60 & \cellcolor{orange!10}82.54 (\textcolor{red}{+4.94})
    & 78.96 & 75.28 & 80.33 & 80.39 & 84.02 & \cellcolor{orange!10}89.43 (\textcolor{red}{+5.41})
    & 83.94 & 82.32 & 86.01 & 82.71 & 86.73 & \cellcolor{orange!10}91.73 (\textcolor{red}{+5.00})\\
$\bullet$ & $\bullet$ & $\bullet$ & $\bullet$ 
    & 70.24 & 69.03 & 71.13 & 77.61 & 79.33 & \cellcolor{orange!10}86.36 (\textcolor{red}{+7.03})
    & 79.48 & 77.71 & 80.86 & 85.78 & 85.31 & \cellcolor{orange!10}91.43 (\textcolor{red}{+6.12})        
    & 84.74 & 88.46 & 89.45 & 89.64 & 90.84 & \cellcolor{orange!10}95.41 (\textcolor{red}{+4.57})\\ 
\bottomrule
\end{tabular}
}\end{center}
\label{tab:ex_2}
\vspace{-0.15in}
\end{table*}

\begin{figure*}[t]
\centering
\begin{minipage}{0.48\textwidth} 
    \centering 
    \includegraphics[width=\textwidth]{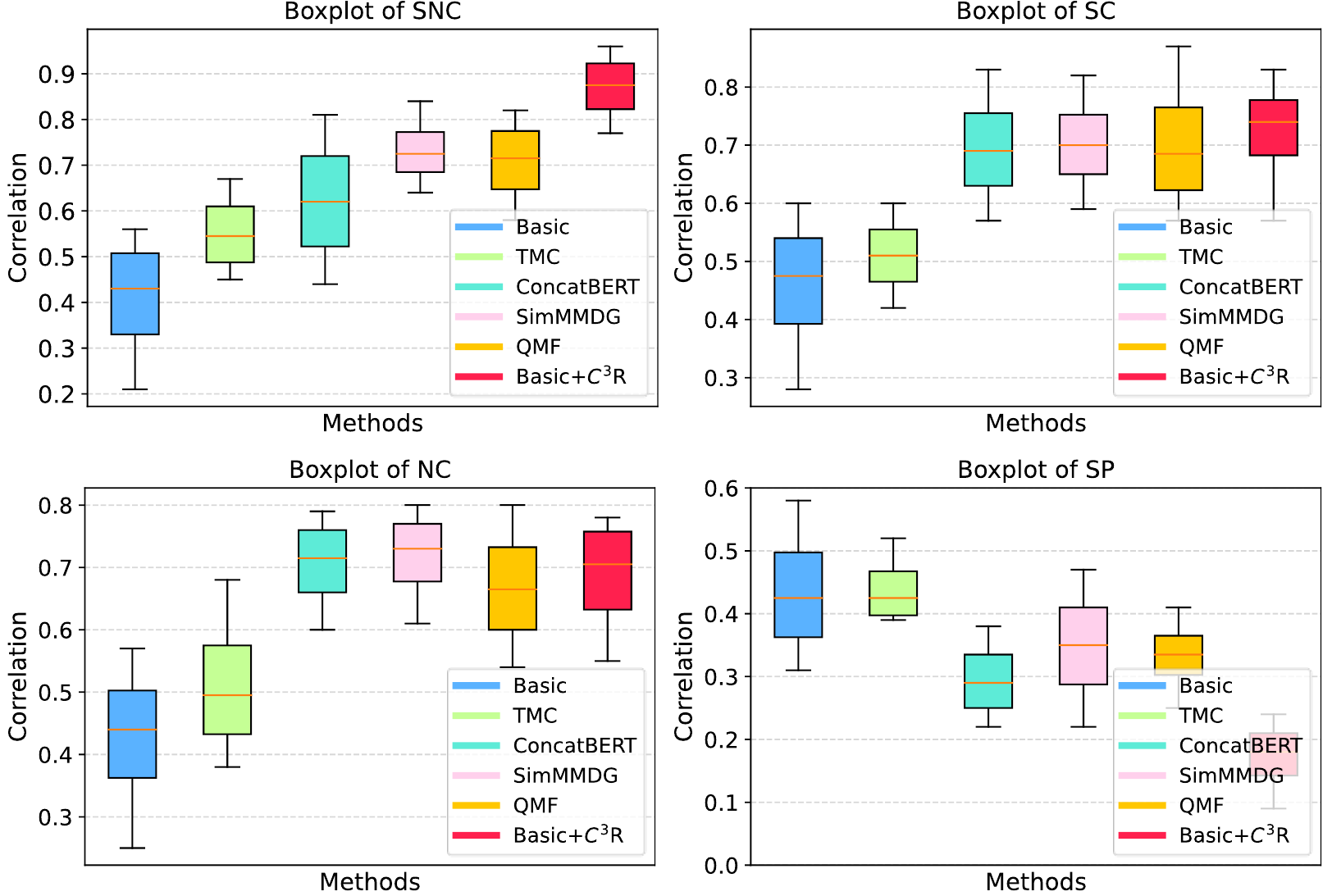}
    \caption{Evaluation for the property of learned representations (SNC, SC, NC, and SP). See \textbf{Appendix H} for details.}
    \label{fig:ex_3}
\end{minipage}
\hfill 
\begin{minipage}{0.48\textwidth} 
    \centering 
    \includegraphics[width=\textwidth]{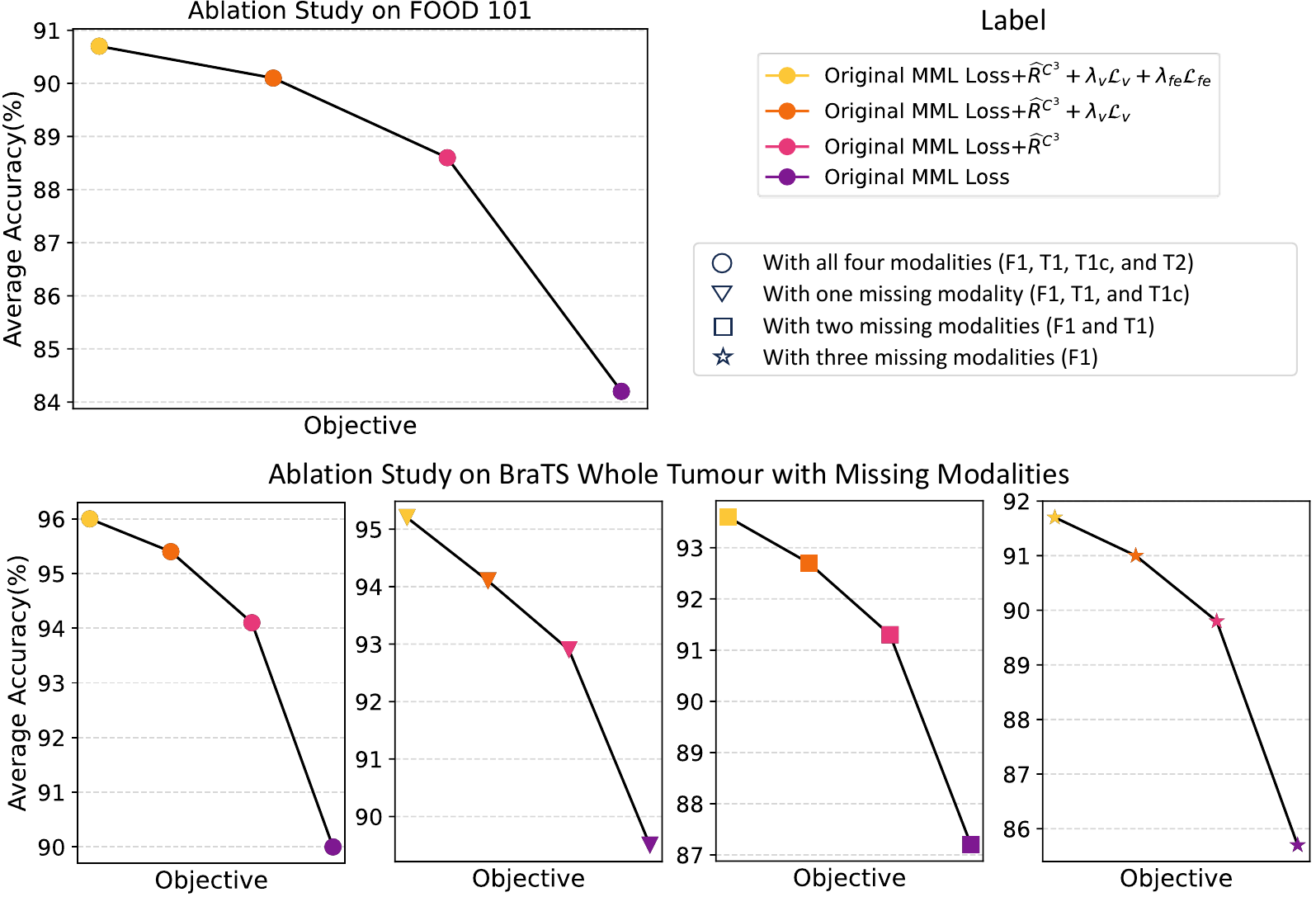}
    \caption{Ablation study of $C^3$R (performance when removing different regular terms). See \textbf{Appendix H} for details.} 
    \label{fig:ex_abla}
\end{minipage}
\end{figure*}

\section{Learning Causal Complete Representations}
\label{sec:6}

Based on the above theoretical results, we propose a plug-and-play method, Causal Complete Cause Regularization ($C^3$R), which is built upon the $C^3$ risk to extract causal complete representations from observable multi-modal data. 

Specifically, we first introduce the constraints of the instrumental variable $V$ (\textbf{Theorem \ref{theorem:variable_Z}}) to make the MML model $f_\theta$ learn causal representations while satisfying the identifiability in practice. Then, we minimize the $C^3$ risk of the learned representations (\textbf{Theorem \ref{theorem:c3r}})
to ensure the causal completeness of the learned representation, i.e., causal sufficiency and necessity. In summary, the objective of $C^3$R is a combination of the above two-step objectives, which can be embedded in various MML models. It can be expressed as: 
\begin{equation}\label{eq:objective_c3r}
\begin{split}
    \underset{f_\theta}{\min}\,\widehat{R}^{C^3}+\lambda_{v}\mathcal{L}_{v}+\lambda_{fe}\mathcal{L}_{fe}
\end{split}
\end{equation}
where $\lambda_v$ and $\lambda_{fe}$ denotes the importance weights, the three terms: (i) \(\widehat{R}^{C^3}\) (Eq.\ref{eq:c3r}) constrains the model to learn a causally complete representation, i.e., one that exhibits minimal sufficiency and necessity risk; (ii) \(\mathcal{L}_{v}\), built upon \textbf{Theorem \ref{theorem:variable_Z}}, employs the instrumental variable \(V\) to guide the model toward the causal representation, corresponding to the real-world branch; (iii) \(\mathcal{L}_{fe}\) stems from the counterfactual construction condition in Eq.\,\ref{eq:twin}, specifically $\mathcal{L}_{fe}=\frac{1}{N}\sum_{i=1}^{N} \mu_{\text{KL}}\bigl(\overline{\text{Z}}_{c,i},\, \widehat{\text{Z}}_{c,i}\bigr)$, ensuring that \(\widehat{\text{Z}}_{c,i}\) remains within the feasible region. Thus, by minimizing the above objective, the $C^3$ score of the learned representation will be higher while maintaining the required conditions for practical implementations. Then, it makes the learned representation causally sufficient and necessary with high confidence.

\section{Experiments}
\label{sec:7}

In this section, we conduct extensive experiments on various benchmark datasets to verify the effectiveness of $C^3$R. More details and experiments are provided in \textbf{Appendix E-H}.

\subsection{Experimental Settings}
\label{sec:7.1}

\textbf{Datasets} We select six datasets: (i) scenes recognition on NYU Depth V2 \cite{silberman2012indoor} and SUN RGBD \cite{song2015sun} with RGB and depth images; (ii) image-text classification on UPMC FOOD101 \cite{wang2015recipe} and MVSA \cite{niu2016sentiment} with image and text; (iii) segmentation considering missing modalities on BraTS \cite{menze2014multimodal,bakas2018identifying} with Flair, T1, T1c, and T2; and (iv) synthetic MMLSynData (see \textbf{Appendix D.5}).

\textbf{Implementation Details} We use a three-layer MLP with activation functions \cite{clevert2015fast} as the representation learner. The hidden vector dimensions of each layer are specified as 64, 32, and 128, while the learned representation is 64. 
For optimization, we employ the Adam optimizer \cite{kingma2014adam} with Momentum and weight decay set at $0.8$ and $10^{-4}$. The initial learning rate is established at 0.1, with the flexibility for linear scaling as required. Additionally, we use grid search to set the hyperparameters $\lambda_v=0.75$ and $\lambda_{fe}=0.4$. All experimental procedures are executed in five runs via NVIDIA RTX A6000 GPUs. Codes can be found in our Github repository\footnote{
Codes can be found in \url{https://github.com/WangJingyao07/Multi-Modal-Base}}.

\subsection{Results}
\label{sec:7.2}

\textbf{Performance and robustness analysis} 
To evaluate the effectiveness of $C^3$R, we record the performance change across various MML baselines after introducing $C^3$R under Gaussian noise (for image modality) and blank noise (for text modality) following \cite{han2020trusted,zhang2023provable,ma2021trustworthy}. We record both the average and worst-case accuracy. The results are shown in \textbf{Table \ref{tab:ex_1}}. We can observe that $C^3$R achieves stable improvements in both the average and worst-case accuracy. This proves the superior effectiveness and robustness of $C^3$R.

\textbf{When faces the problem of missing modalities} 
Considering the missing modality issues, we evaluate the performance of $C^3$R and several strong baselines \cite{wang2023learnable,bakas2018identifying,zhang2022mmformer} on all 15 possible combinations of missing modalities on BraTS. From \textbf{Table \ref{tab:ex_2}}, we can observe that (i) $C^3$R brings significant performance improvements; (ii) $C^3$R can reduce the learning gap for the representations on different modal semantics, i.e., reducing the accuracy gap of learning on the difficult-to-identify Fl and T1 modalities and the easy T1c. This demonstrates the superiority of $C^3$R and the advantage of causally complete representation in missing modality issues.

\textbf{Learning causal complete representations} 
To evaluate $C^3$R's ability to extract causal complete causes, we conduct experiments: (i) construct four types of MML data, i.e., sufficient and necessary (SNC), sufficient but unnecessary (SC), necessary but insufficient causes (NC), and spurious correlations (SP) following \cite{yang2024invariant} (see \textbf{Appendix H} for more details); then (ii) evaluate their correlation with the learned representation based on the distance metric \citep{jones1995fitness}, i.e., the higher the score, the stronger the correlation. The results are shown in \textbf{Figure \ref{fig:ex_3}} (see \textbf{Appendix D} and \textbf{H} for details). The representations learned by $C^3$R have a higher correlation with SNC and lower with SP. This proves that $C^3$R can learn causal complete representations more effectively, while other methods are hard to achieve.

\textbf{Ablation Study}
To evaluate the effect of each item in $C^3$R (Eq.\ref{eq:objective_c3r}), we evaluate the performance change on LCKD+$C^3$R. We record the performance changes before and after the model introduced specific terms. We consider the standard and corner case (missing modality), and conduct experiments on FOOD 101 and BraTS. The results are shown in \textbf{Figure \ref{fig:ex_abla}}. We can observe that each item plays an important role. See \textbf{Appendix H} for more experiments, including the analyses about model efficiency and parameter sensitivity. We provide a brief summary about all the experiments conducted in this paper, as shown in Table 3.


\section{Related Work}
\label{sec:2}

Multi-modal learning aims to learn good representations through multiple modalities for accurate prediction. Recently, multiple methods \cite{mmlsurvey,jia2022multi,wang2023amsa,fan2024improving} have been proposed to solve MML tasks through modality consistency, e.g., tokenize diverse modalities into sequences and utilize Transformers for joint learning \cite{bao2022vlmo, wang2022ofa}, whereas CLIP \cite{radford2021learning,fan2024improving}, ALIGN \cite{jia2021scaling}, etc. employ distinct encoders for each modality and utilize contrastive loss to synchronize features. These methods align features from different modalities into the same space to capture the primary events. Another type of work \cite{jiang2023understanding,dong2024simmmdg,wang2023multi,wu2020modality} proposed to additionally focus on modality specificity. They divide the MML features into modality-shared and modality-specific components and learn separately. However, all these methods may result in learning insufficient or unnecessary information. Meanwhile, they rely on strong assumptions like semantic alignment \cite{implicit1,wang2023amsa,explicitly1,explicitly2} and may be limited to ideal data \cite{ge2023improving,zhang2023generalization}, affecting model performance.
For causal sufficiency and necessity, \cite{pearl2009causality} first proposed the related concepts, and \cite{yang2024invariant} applied it to domain generalization with assumptions in an ideal environment. The differences between our methodology and previous causal-related works include problem settings, theorems, implementations, learning objectives, etc. (More comparison and analyses are provided in \textbf{Appendix D.6}).
We explore the MML-specific concepts of causal completeness without strong assumptions, ensuring the effectiveness of the learned representations.


\section{Conclusion}
\label{sec:8}
We explore the MML-specific concepts of causal sufficient and necessary without exogeneity and monotonicity assumptions. To measure whether the learned MML representation is causally complete, we present the definition, identifiability, and measurement of $C^3$ with theoretical support. Based on these results, we propose a plug-and-play method $C^3$R to promote MML learning causal complete representations. Extensive experiments demonstrate its effectiveness. 

\section*{Impact Statement}
This paper presents work whose goal is to advance the field of 
Machine Learning and Multi-Modal Learning. There are many potential societal consequences 
of our work, none of which we feel must be specifically highlighted here.

\section*{Acknowledgements}
The authors would like to thank the anonymous reviewers for their valuable comments. This work was supported in part by the National Natural Science Foundation of China (Grant No. 62406313), in part by the National Key R\&D Program of China (Grant No.2023YFF0725001), in part by the National Natural Science Foundation of China (Grant No.92370204), in part by the guangdong Basic and Applied Basic Research Foundation (Grant No.2023B1515120057) in part by Guangzhou-HKUST (GZ) Joint Funding Program (Grant No.2023A03J0008), Education Bureau of Guangzhou Municipality.

\bibliography{reference}
\bibliographystyle{icml2025}

\end{document}